\documentclass{article}

\usepackage[preprint]{amlc_2022}

\usepackage[utf8]{inputenc} 
\usepackage[T1]{fontenc}    
\usepackage{hyperref}       
\usepackage{url}            
\usepackage{booktabs}       
\usepackage{amsfonts}       
\usepackage{nicefrac}       
\usepackage{microtype}      
\usepackage{xcolor}         
\usepackage{graphicx}

\newcommand{\ours}[0]{{RTVAE-Multi}}
\newcommand{\ourssingle}[0]{{RTVAE-Single}}
\newcommand{\baseline}[0]{{ACTOR-Multi}}
\newcommand{\baselinesingle}[0]{{ACTOR}}
\newcommand{\myparagraph}[1]{\vspace{0.1em}\noindent\textbf{#1}}

\usepackage{adjustbox}
\usepackage{bm}  
\usepackage{subcaption}
\usepackage{caption}
\usepackage{makecell}
\usepackage{booktabs}
\usepackage{multirow}
\usepackage[numbers]{natbib}

\usepackage[symbol]{footmisc}

\newcommand{\R}{\mathbb{R}}

\newcommand{\floor}[1]{\lfloor #1 \rfloor}

\title{Recurrent Transformer Variational Autoencoders for \\ Multi-Action Motion Synthesis}

\author{
Rania Briq$^{1,2,\ast}$ \qquad
Chuhang Zou$^{2}$ \qquad
Leonid Pishchulin$^{2}$ \qquad
\\
\textbf{Chris Broaddus}$^{2}$ \qquad
\textbf{J\"urgen Gall}$^{1}$
\\
$^1$University of Bonn\qquad
$^2$Amazon
}

\begin{document}

\maketitle

\begin{abstract}
We consider the problem of synthesizing multi-action human motion sequences of arbitrary lengths. Existing approaches have mastered motion sequence generation in single action scenarios, but fail to generalize to multi-action and arbitrary-length sequences. We fill this gap by proposing a novel efficient approach that leverages expressiveness of Recurrent Transformers and generative richness of conditional Variational Autoencoders. The proposed iterative approach is able to generate smooth and realistic human motion sequences with an arbitrary number of actions and frames while doing so in linear space and time.
We train and evaluate the proposed approach on PROX and Charades datasets~\cite{PROX:2019,sigurdsson2018charadesego}, where we augment PROX with ground-truth action labels and Charades with human mesh annotations. Experimental evaluation shows significant improvements in FID score and semantic consistency metrics compared to the state-of-the-art.
\end{abstract}
\footnotetext{$\ast$ Work done during internship at Amazon.}

\section{Introduction}
\label{sec:intro}
This paper addresses the problem of synthesizing multi-action human motion sequences of arbitrary lengths. This is a major research problem with a vast range of applications, including animating avatars and virtual characters~\cite{lee2002interactive,lee2006precomputing,holden2017phase,starke2019neural,won2020scalable,2018-TOG-SFV, won2020scalable}, character visualization in stories~\cite{li2019storygan}, inpainting of missing frames in videos and completion of action sequences~\cite{wexler2004space,cai2018deep}, and synthesizing realistic motion under 3D scene constraints~\cite{hassan2021stochastic,wang2021synthesizing}. Generating realistic motions of humans performing multiple action over longer periods of time is a challenging task, since it requires modeling dependencies between multiple possibly overlapping actions in the temporal domain, while ensuring that the transition points between different actions are continuous and smooth. Recent approaches \cite{guo2020action2motion,petrovich2021action} were able to considerably improve the fidelity of synthesized motion by building on conditional Variational Autoencoders (VAEs)~\cite{kingma2013auto} and Transformers~\cite{vaswani2017attention}. However, both approaches are restricted to generating single action motion sequences only. Furthermore, as motion sequences are generated in a single shot in a non-causal way, their generation is typically limited to short sequences of a pre-defined length to avoid running out of memory, which often results in context fragmentation. On the other hand, Recurrent Transformers (RT)~\cite{choromanski2020rethinking,child2019generating,roy2021efficient,dai2019transformer} have been shown to combine the expressiveness of self-attention mechanism~\cite{vaswani2017attention} to model long-range dependencies with the efficiency of recurrent architectures~\cite{hochreiter1997long}. However, RTs lack conditioning and thus have not been used for motion sequence synthesis.

\begin{figure}[t]
	\centering
	\includegraphics[width=10.0cm]{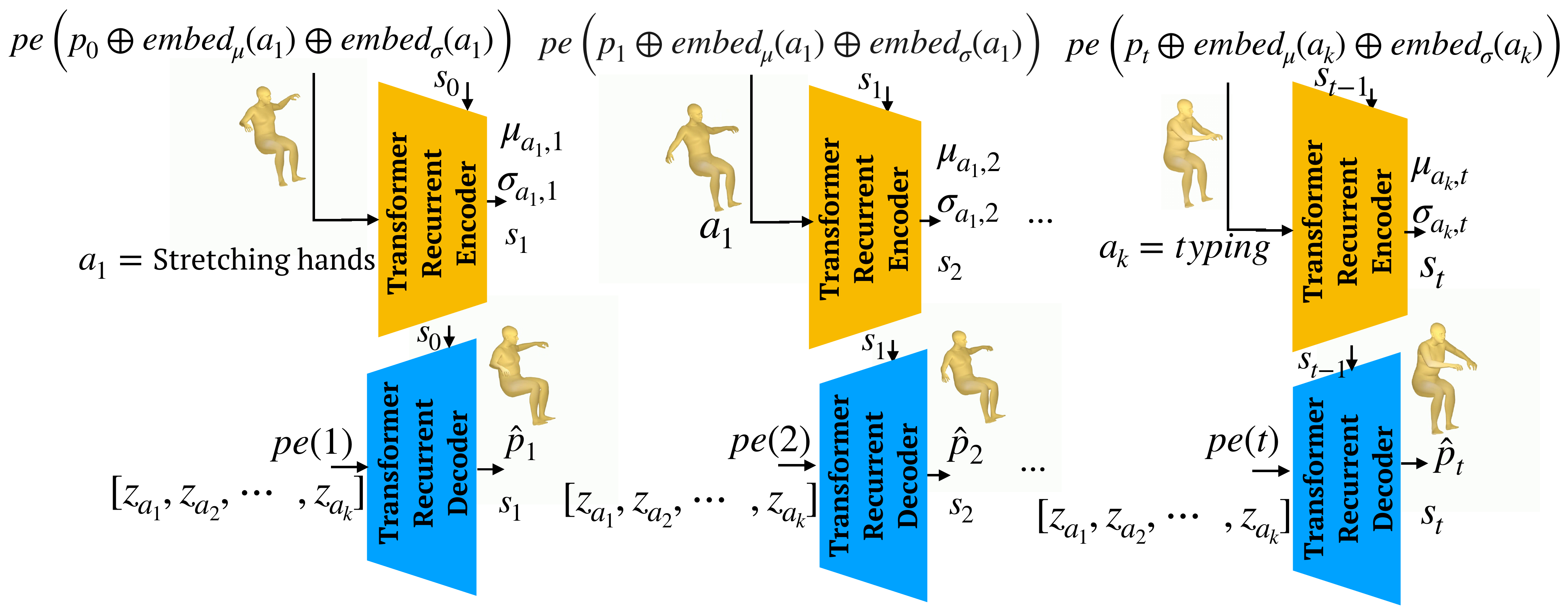}
	\caption{Overview of our approach. The encoder and decoder steps are unrolled to demonstrate the synthesis process of sequences conditioned on multiple actions. At iteration $t$, the input to the encoder is a SMPL pose denoted by $p_t$, concatenated with the embedding of the input action $a_i$, where $a_i$ spans multiple frames, and \textit{pe} is the positional embedding. The previous hidden state used to compute the output at $t$ is denoted by $s_{t-1}$. For each action, we save the VAE distribution parameters $\mu_{a_i,t_i}$ and $\sigma_{a_i,t_i}$ where $t_i$ denotes the end timestamp of action $a_i$. Once the encoder has seen all frames, we sample $k$ latent vectors $z_i\sim N(\mu_{a_i,t_i},\sigma_{a_i,t_i})$ each corresponding to the entire subsequence matching $a_i$. The decoder transformer query sequence corresponds to the positional embedding of timestamp $t$, and $[z_{a_1},z_{a_2},\cdots\ , z_{a_k}]$ is a stacking of all the latent vectors samples from encoder output. The decoder is optimized to output a reconstruction of the input poses using an MSE loss term.}
	\label{fig:approach_overview}
\end{figure}

We propose a novel approach that takes an ordered set of action labels as input and synthesizes an arbitrary-length motion sequence of 3D human meshes performing these multiple potentially-overlapping actions. At the core of our approach lies a novel spatio-temporal formulation that marries Recurrent Transformers (RT) with conditional Variational Autoencoders (VAE) to generate multi-action sequences. We thus dub the proposed approach as \ours. The approach (cf. Fig.~\ref{fig:approach_overview}) is trained using sequences of SMPL~\cite{SMPL-X:2019,SMPL:2015} 3D body meshes of people performing multiple actions, and associated ordered sets of action labels. During training, at each timestamp, we concatenate a 3D body mesh with the input actions embeddings and feed them into the encoder together with the previous hidden state. Once the encoder has seen all frames, we sample multiple latent vectors from the distribution parameters that were predicted by the encoder, each corresponding to the whole action sub-sequence. We then add the corresponding action class embedding to further distinguish these vectors in the latent space. In the decoding phase, the decoder uses the query sequence corresponding to the positional embedding of a given timestamp and the stacked latent vectors to reconstruct the input poses. During inference, the length of sequences that can be generated is arbitrary, and each frame is generated at constant time and memory.

To train \ours~ we require a dataset with both multi-action sequences annotated with GT action labels, and frame-wise SMPL~\cite{SMPL-X:2019,SMPL:2015} 3D body mesh fittings. Since no such dataset is publicly available, we found PROX dataset~\cite{PROX:2019} to be closest to our requirements, and have therefore augmented it with GT multi-action labels. Additionally, we experiment with Charades dataset which was captured in unconstrained cluttered environments. 

Our main contribution is a novel spatio-temporal formulation that combines the expressiveness and efficiency of Recurrent Transformers with generative richness of conditional Variational Autoencoders to generate arbitrary-length multi-action motion sequences. In contrast to previous works~\cite{guo2020action2motion,petrovich2021action}, we address a more challenging problem of multi-action motion synthesis, are able to generalize to arbitrary number of actions per generated sequence, and to generate arbitrary length sequences where space and time requirements grow linearly in the number of frames. In Sec.~\ref{sec:experiments} we demonstrate significant improvements in motion synthesis over multi-action extension of~\cite{petrovich2021action}, while performing on-par with~\cite{petrovich2021action} in single-action scenarios. 

Our second and third contributions is repurposing two publicly available datasets to the proposed task by extending their annotations. We augment PROX dataset~\cite{PROX:2019} with GT multi-action labels, thus completing the dataset to contain multi-action motion sequences, per-frame SMPL~\cite{SMPL-X:2019,SMPL:2015} fittings \emph{and} GT action labels. We also augment Charades dataset ~\cite{sigurdsson2018charadesego} with SMPL fittings where we propose a method to obtain SMPL meshes in continuous frames from action recognition datatsets while detecting and filtering out bad estimates from sequences. 

Our fourth contribution is a thorough experimental evaluation on the challenging task of synthesizing arbitrary-length motion sequences with an arbitrary number of actions and comparisons to the state-of-the-art and strong baselines. In Sec.~\ref{sec:experiments} we demonstrate that the proposed design choices result in significant improvements in action recognition when evaluating a pre-trained off-the-shelf action recognition approach~\cite{yu2017spatio} on generated sequences.

\section{Related Work}
\label{sec:related_work}
Motion synthesis is a widely researched problem in graphics and computer vision \cite{li2017auto,guo2020action2motion,wang2021synthesizing,petrovich2021action,yan2019convolutional,barsoum2017hpgan}. 
\cite{martinez2017human,li2017auto, butepage2017deep} condition a future sequence on a few past frames and rely on autoregressive models such as LSTMs~\cite{hochreiter1997long}. These models are inherently incapable of modeling long-term dependencies between frames and their ability to generate realistic motion degenerates for longer sequences. Works such as \cite{diba2019dynamonet} jointly predict future frames in action classification frameworks in order to learn action-specific motion representation. Deep generative models such as Variational Autoencoders (VAE)~\cite{kingma2013auto} and Generative Adversarial Networks (GANs)~\cite{goodfellow2014generative} have made breakthroughs in synthesizing high-fidelity visual content such as images, videos and human motion~\cite{reed2016generative,tao2020df,li2022stylet2i}. For example, Barsoum etal ~\cite{barsoum2017hpgan} proposes a GAN-based Seq2seq model while~\cite{yan2019convolutional} proposes a GAN-based model using graph convolutional network that samples a latent vector from an observed prior. Methods based on VAEs for motion synthesis conditioned on action labels have been proposed in~\cite{guo2020action2motion}, while~\cite{petrovich2021action} further proposed generating the full body mesh rather than the 3D pose. Other tasks include motion transfer or video completion in which generating human videos is guided by motion sequences that are generated based on a semantic label \cite{yang2018pose,cai2018deep}. Similarly, controllable characters using user-defined controls are used to generate an image sequence of a given person~\cite{gafni2019vid2game}, or to animate characters and avatars~\cite{lee2006precomputing,starke2019neural,won2020scalable}. In a related task~\cite{li2019storygan} visualize a given story and its characters by generating a sequence of images at the sentence level and~\cite{ahuja2019language2pose} synthesize motion sequences conditioned on natural language. Another group of approaches predicts human motion constrained by scene context or human-object interaction~\cite{wang2021synthesizing,hassan2021stochastic,corona2020context,cao2020long}.

\myparagraph{Variational Autoencoder Transformer.} \cite{petrovich2021action} proposed the currently state-of-the-art approach to synthesizes motion sequences conditioned on a given action label. The underlying generative method is a Transformer-based~\cite{vaswani2017attention} VAE~\cite{kingma2013auto} which learns the latent distribution of the observed pose sequences. \cite{petrovich2021action} is limited to single action sequence generation, where each sequence is generated in one shot in a non-causal way. This bounds the length of the sequence by the resources available on a machine, a limitation further amplified by quadratic complexity attention computation, which requires splitting long sequences into several parts thereby causing context fragmentation.

\myparagraph{Recurrent Transformers.}
Transformer~\cite{vaswani2017attention} is an expressive network architecture that relies on the self-attention mechanism for weighing different elements of the input. Several approaches have been proposed to alleviate the quadratic complexity needed to compute the attention~\cite{choromanski2020rethinking,child2019generating,roy2021efficient,dai2019transformer}. One prominent work is a recurrent approach based on formulating the attention using kernel functions~\cite{katharopoulos2020transformers}, making it possible to calculate the attention in linear time. 

\section{Multi-action Recurrent Transformer Variational Autoencoder}

Given an ordered set of action labels $A=\{a_1, a_2, ..., a_k\}$, we propose a model that synthesizes a motion sequence of 3D human meshes performing these actions. The ability to generate long sequences that are plausible and coherent requires allowing information to flow persistently across the temporal domain. To that end, we propose using a recurrent model with a hidden state that maintains compact yet long-range information to help it reason about the sequence.

\subsection{Approach} 
In order to overcome the aforementioned limitations of~\cite{petrovich2021action}, we propose to extend their approach to generate sequences of an arbitrary length conditioned on an arbitrary number of action labels. An overview of the approach is shown in Fig.~\ref{fig:approach_overview}. The input during training is a sequence consisting of multiple possibly overlapping actions and an ordered set of corresponding action labels. The transformer network additionally receives its last hidden state as part of the input, and outputs an encoding of the input representing the current pose and distribution parameters $\mu$ and $\sigma$. We save the distribution parameters corresponding to the last frame of every action, since we hypothesize that the encoder learns a better representation as it accumulates more knowledge of an action sub-sequence. We then sample $k$ latent vectors $z_i\in\R^d$ each corresponding to a sub-sequence of one action. In order to guarantee a continuous and uniform representation for the entire sequence, especially at the cross points where action $a_i$ ends and $a_{i+1}$ starts, we stack the latent vectors of all actions. In the decoding phase, we do not rely on timestamps to decide the number of frames synthesized for each action, and instead, we let the decoder reason about this number. This makes it possible to train the decoder in a weakly supervised manner, as it learns to relate an action-specific latent vector with a generated sub-sequence, but that also entails a limitation where during inference, it is not straightforward to specify the length of the sub-sequence corresponding to an action.

\subsection{Data pre-processing}
\label{sec:data_preprocessing}
\begin{figure}[h]
	\setlength\tabcolsep{4pt}
		\begin{tabular}{llllll}
 {\includegraphics[height=4cm, width=3cm]{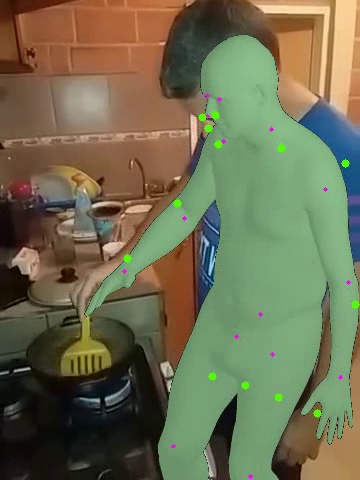}} &{\includegraphics[height=4cm, width=4cm]{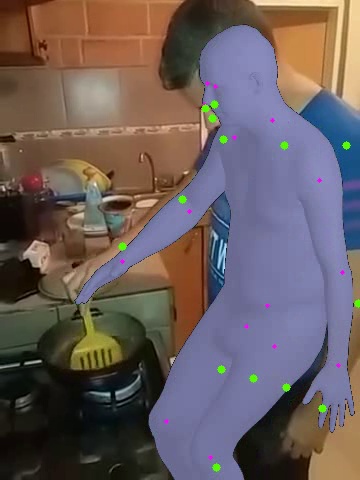}} &{\includegraphics[height=4cm, width=3cm]{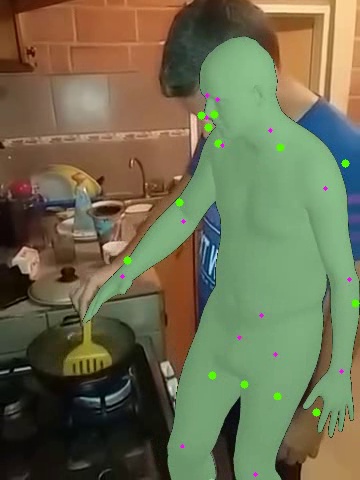}}
	\end{tabular}
	\caption{Example of a video from Charades dataset \cite{sigurdsson2018charadesego}. The larger green dots are the predicted 2D keypoints, and the pink dots are the projection of the SMPL 3D joints into 2D. It can be seen that in the middle frame there is some deviation in the right-side hand keypoint, therefore it is detected as a bad-quality prediction and is filtered out. However, the keypoints in the preceding and subsequent frames match up to a small deviation.}
	\label{fig:preprocessing}
\end{figure}

 Since we are interested in making our approach relevant for any action recognition dataset which already includes action annotations but leave out the 3D human body mesh, we present a method that makes it possible to train the proposed approach over such datasets. Directly applying frameworks that predict these meshes especially in cluttered or blurred scenes can result in sequences with bad-quality estimates that can divert the model. Since we do not want to discard an entire sequence every time there are a few frames with bad estimates, we propose a method to detect and filter out these meshes by relying on 2D keypoints estimates. 2D keypoints are generally easier to estimate and we obtain better-quality results in the 2D space. If a 3D SMPL joint estimate is good, then its 2D projection should not deviate from the 2D estimate by a large margin and its confidence in prediction is high. We therefore project the 3D SMPL joints into the 2D space and compare the scaled Euclidean distance between each two matching joints, i.e.\, $\frac{\pi(\hat{p}^{i}_{3D})-\hat{p}^{i}_{2D}}{s}\leq \tau$, where $\pi$ indicates the projection operation and $s$ the scaling factor proportional to the head size. The number of keypoints with a deviation larger than $\tau=1$ should be kept low ($10$ in our experiments). We obtain 2D keypoints using OpenPose\cite{8765346}. If a sequence contains only a few frames with bad estimates, we divide up the sequence around these frames and we rely on the original action timestamps to match up the action labels to these subsequences. An illustration of the technique is shown in Fig.~\ref{fig:preprocessing}.

\section{Experiments}
\label{sec:experiments}
\myparagraph{Datasets.} Existing motion datasets such as HumanAct12~\cite{guo2020action2motion} and Human3.6M~\cite{h36m_pami} consist of single-action sequences, making it hard to train and evaluate the proposed model. We thus extended two datasets to include both action and SMPL annotations.

\myparagraph{PROX dataset.}
 PROX dataset~\cite{PROX:2019} originally contains 3D pose annotations and SMPL~\cite{SMPL-X:2019,SMPL:2015} fittings, and we augment it to fulfill our needs by adding GT action labels. We identified $50$ action labels in $60$ various recorded scenarios in $12$ different 3D scenes in which 20 subjects are interacting with the scene. The number of frames is 100K and we split the dataset into training and test sets, where the training set consists of $27$ scenes with subjects performing actions in indoor environments such as in the library, cafeteria and office. The test set consists of $3$ scenes, some of which have only ground truth (GT) action annotations without SMPL fittings. The scenes consist of $1000-3000$ frames and the actions performed can be intricate and are not repetitive or simple in nature. Example actions include \textit{writing on board},\textit{crossing legs}, and \textit{dancing}.
 
 \begin{table*}[ht]
	\centering
	\begin{adjustbox}{width=1.0\textwidth}
	\small
	\begin{tabular}{lccccccccccc}
		\hline
		  seq. length&method &accuracy\textsubscript{tr\textsubscript{gt}}& accuracy\textsubscript{tr\textsubscript{gen}}$\uparrow$&accuracy\textsubscript{test\textsubscript{gen}}$\uparrow$&FID\textsubscript{tr\textsubscript{gen}}$\downarrow$&diversity\textsubscript{tr\textsubscript{gt}}&multimodality\textsubscript{tr\textsubscript{gt}}&diversity\textsubscript{tr\textsubscript{gen}}&multimodality\textsubscript{tr\textsubscript{gen}}\\
		  \hline
		\multirow{2}{*}{60} &\baseline&\multirow{2}{*}{$98.1$}& $74.3\pm1.26$&$60.0\pm6.1$&$680\pm11.46$&\multirow{2}{*}{$86.57$}&\multirow{2}{*}{$26.96$}&{$65.67\pm0.51$}&$23.66\pm0.46$\\
		&\ours&&$\textbf{82.8}\pm0.84$&$\textbf{61.8}\pm2.62 $&$\textbf{590}\pm6.23$&&&$73.0\pm0.39$&$20.5\pm0.36$\\
		\hline
		\multirow{2}{*}{80} &\baseline&\multirow{2}{*}{$97.1$}&$56.2\pm0.92$&$41.5\pm1.54$&$1197.4\pm33.43$&\multirow{2}{*}{89.2}&\multirow{2}{*}{35.16}&$67.58\pm1.32$&$26.69\pm0.35$\\
		&\ours& &$\textbf{74.3}\pm1.11$&$\textbf{49.4}\pm1.72 $&$\textbf{996}\pm23.12$&&&$75.5\pm0.64$&$25.85\pm0.39$\\
		\hline
		\multirow{2}{*}{120} & \baseline&\multirow{2}{*}{$99.1$}&$38.3\pm0.47$&$27.9\pm3.07$&$2938.44\pm35.6$&\multirow{2}{*}{$130.47$}&\multirow{2}{*}{$47.86$}&$97.24\pm0.82$&$38.95\pm0.51$\\
		&\ours& &$\textbf{50.8}\pm0.71$&$\textbf{37.3}\pm1.49$&$\textbf{2584}\pm9.7$&&&$100.14\pm0.8$&$36.81\pm0.54$
	\end{tabular}
	\end{adjustbox}
	\caption{Evaluation on PROX dataset with a maximum of $8$ actions/sequence, while varying sequence length.} 
	\label{tab:eval_increase_seq_multi}
\end{table*} 

\begin{table*}[ht]
	\centering
	\begin{adjustbox}{width=1.0\textwidth}
	\small
	\begin{tabular}{lccccccccccc}
		\hline
		  seq. length&method &accuracy\textsubscript{tr\textsubscript{gt}}& accuracy\textsubscript{tr\textsubscript{gen}}$\uparrow$&accuracy\textsubscript{test\textsubscript{gen}}$\uparrow$&FID\textsubscript{tr\textsubscript{gen}}$\downarrow$&diversity\textsubscript{tr\textsubscript{gt}}&multimodality\textsubscript{tr\textsubscript{gt}}&diversity\textsubscript{tr\textsubscript{gen}}&multimodality\textsubscript{tr\textsubscript{gen}}\\
		  \hline
		\multirow{2}{*}{60} &\baselinesingle~\cite{petrovich2021action}&\multirow{2}{*}{$97.3$}& $\textbf{74.4}\pm\textbf{0.69}$&$\mathbf{65.8\pm3.1}$&$733\pm6.85$&\multirow{2}{*}{$61.64$}&\multirow{2}{*}{$19.14$}&{$49.0\pm0.45$}&$14.14\pm0.38$\\
		&\ourssingle&&$72.7\pm0.75$&$64.8\pm2.34$&$\mathbf{725\pm5.56}$&&&$48.4\pm0.42$&$13.72\pm0.3$\\
		\hline
		\multirow{2}{*}{80} &\baselinesingle~\cite{petrovich2021action}&\multirow{2}{*}{$97.8$}&$70.7\pm1.0$&$61.0\pm1.98$&$1696\pm5.46$&\multirow{2}{*}{77.19}&\multirow{2}{*}{26.33}&$57.72\pm0.37$&$26.33\pm0.28$\\
		&\ourssingle& &$\textbf{71.1}\pm\textbf{0.62}$&$\mathbf{61.5\pm2.87}$&$\mathbf{1631\pm9.84}$&&&$58.67\pm0.52$&$17.77\pm0.34$\\
		\hline
		\multirow{2}{*}{120} & \baselinesingle~\cite{petrovich2021action}&\multirow{2}{*}{$98.4$}&$64.2\pm0.53$&$58.2\pm2.5$&$5311\pm14.73.6$&\multirow{2}{*}{$105.22$}&\multirow{2}{*}{$39.31$}&$81.84\pm0.63$&$25.16\pm0.57$\\
		&\ourssingle& &$\mathbf{65.6\pm1.08}$&$58.2\pm2.69$&$\mathbf{4885\pm23.5}$&&&$81.75\pm0.82$&$25.34\pm0.56$
	\end{tabular}
	\end{adjustbox}
	\caption{Evaluation on PROX dataset when using single action/sequence, while varying sequence length.} 
	\label{tab:eval_increase_seq_single}
\end{table*} 

\begin{figure}
\centering
\setlength\tabcolsep{1pt}
\begin{tabular}{ccc}
 \includegraphics[width=0.35\linewidth]{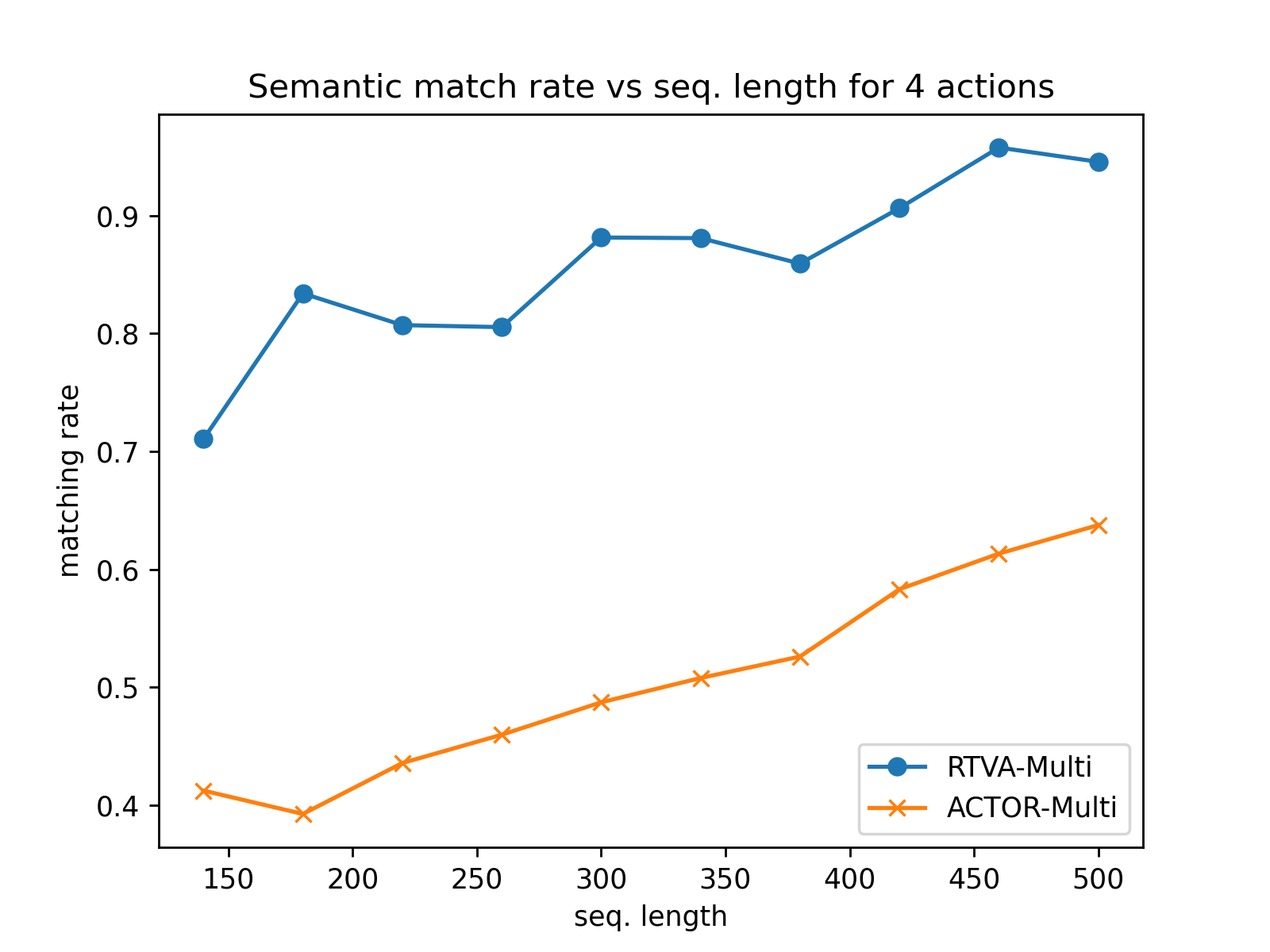}
 & \includegraphics[width=0.35\linewidth]{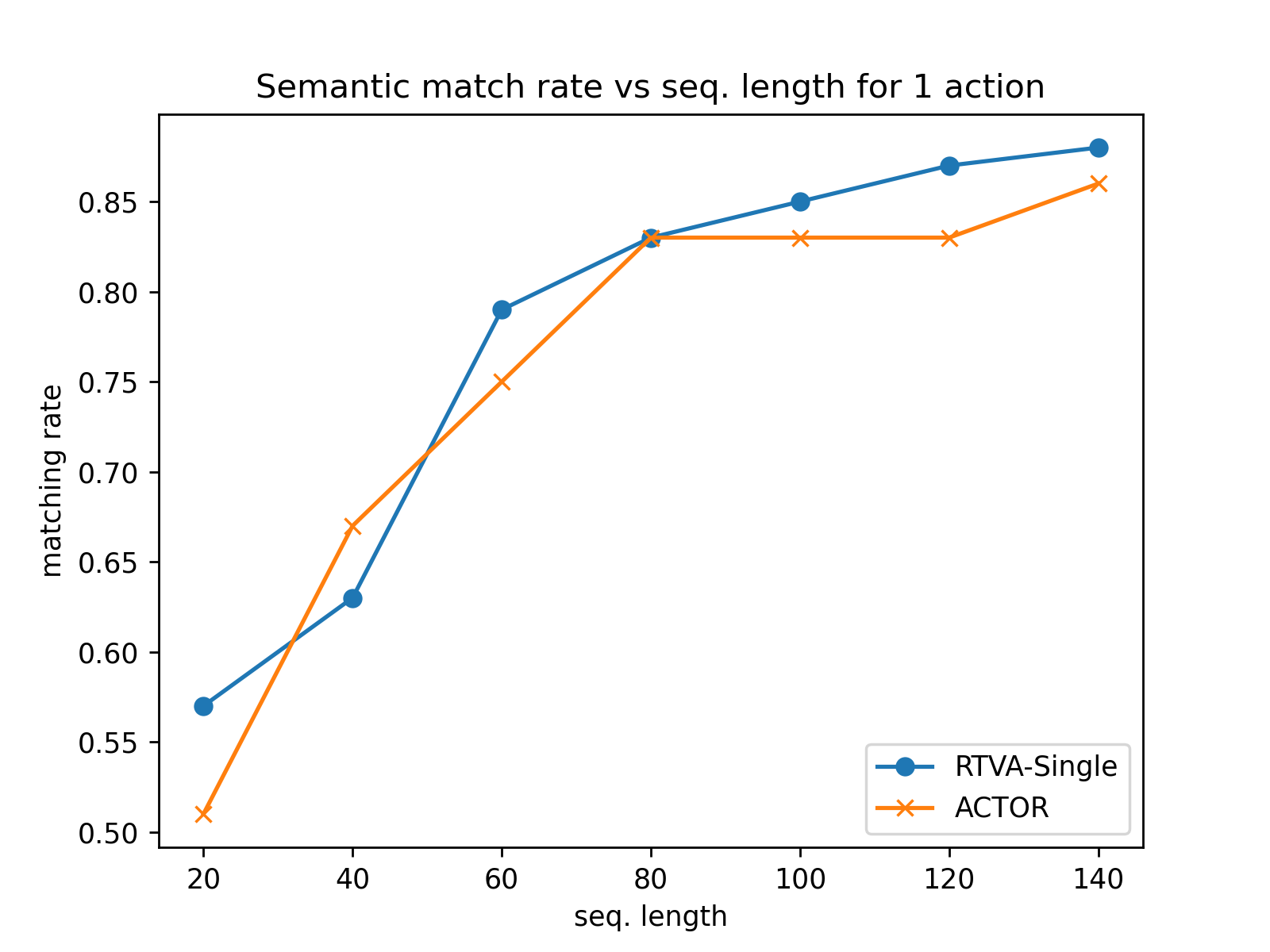}&
 \includegraphics[width=0.35\linewidth]{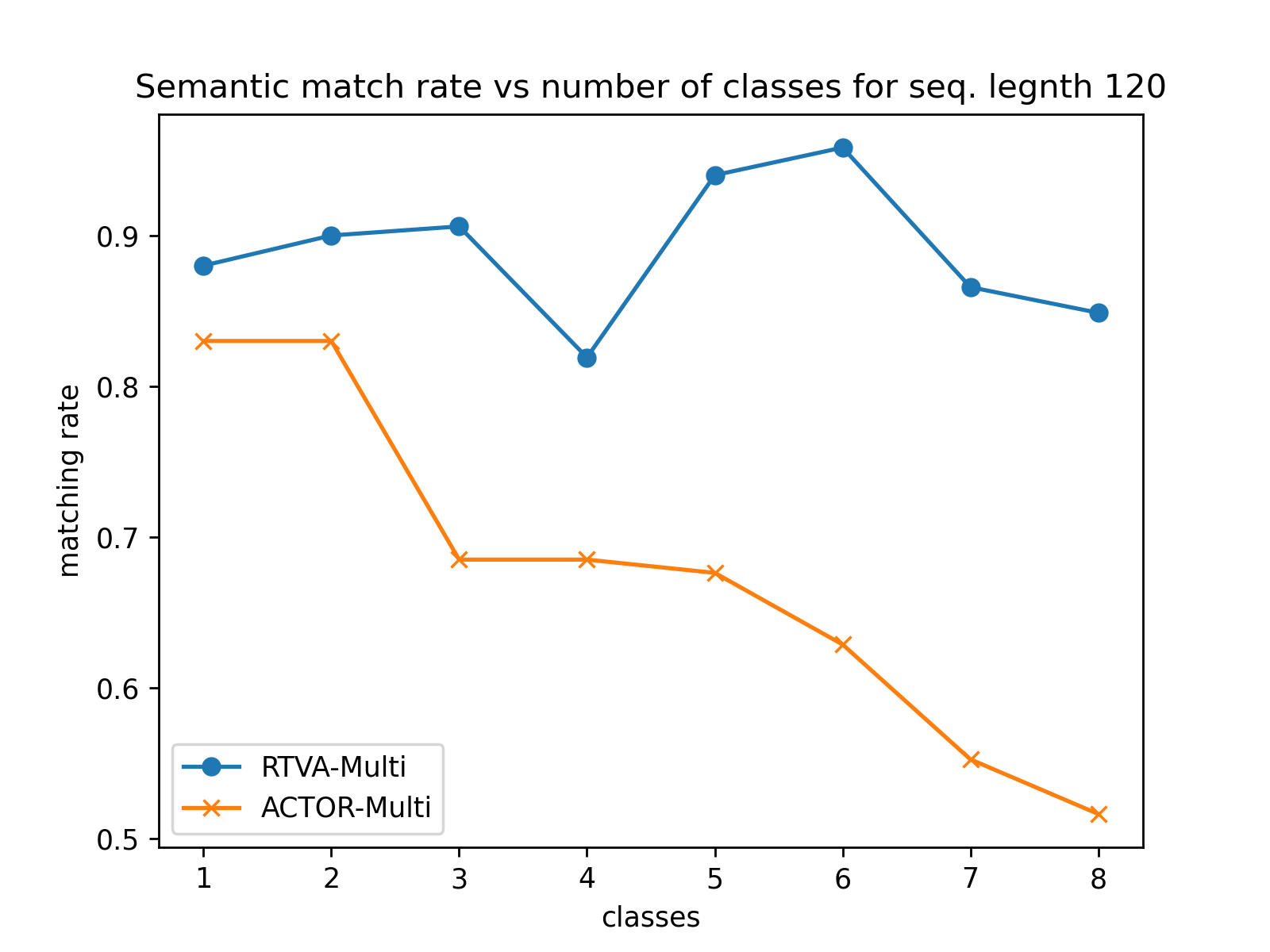} \\
  (a) & (b) &(c)
\end{tabular}
\caption{Match rate of action labels between synthesized and ground-truth samples while varying the sequence length for single action (a) and four actions (b), and varying number of actions for constant sequence length (c).}
\label{fig:all_match_rate}
\end{figure}

\begin{table}[ht]
	\centering
	\begin{adjustbox}{width=0.5\textwidth}
	\small
	\begin{tabular}{lccc}
		\hline
		  method &accuracy\textsubscript{tr\textsubscript{gen}}$\uparrow$&accuracy\textsubscript{test\textsubscript{gen}}$\uparrow$&FID\textsubscript{tr\textsubscript{gen}}$\downarrow$\\
		  \hline
		\ours&$\mathbf{74.3\pm1.11}$&$\mathbf{49.4\pm1.72}$&$\mathbf{996\pm23.12}$\\
		Average stats&$72.8\pm0.47$&$46.9\pm1.98$&$1129\pm9.56$ \\
		All diff-latent&$62.0\pm0.4$&$38.1\pm1.35$&$1249.86\pm4.54 $\\
		Single latent&$55.3\pm0.77$&$43.7\pm2.14$&$1231.73\pm11.67$\\
		W/o look-b.-a.&$29.1\pm1.36$&$16.5\pm2.0$&$2161.42\pm21.46$ 
	\end{tabular}
	\end{adjustbox}
	\caption{Ablative evaluations on PROX using \textit{standard metrics} (max $80$ frames- and max $8$ actions per sequence).} 
	\label{tab:eval_ablative}
\end{table}
 
\myparagraph{Charades dataset.}
In order to check our approach's generalizability, we further evaluate it over Charades-Ego dataset \cite{sigurdsson2018charadesego}, an unstructured action recognition dataset collected through Amazon Mechanical Turk and had been shot from a first and third person perspective. Charades-Ego contains 7860 videos of daily indoors activities with 157 defined actions labels. Its annotations include action classes with timestamps but lack SMPL fittings. We therefore obtain these SMPL fittings and the updated action timestamps based on the technique that we presented in sec.~\ref{sec:data_preprocessing}. When dividing a sequence due to bad SMPL meshes, we discard sequences with less than $30$ frames. There are $1918$ sequences on average and every sequence comprises on average $10$ action classes. Since in this dataset the action classes are imbalanced, to balance the class distribution during training, each sequence is sampled with a probability that is inversely proportional to its action label occurrence in the dataset. We found that this sampling leads to better convergence.

\myparagraph {Baseline.}
Since we are the first to address multi-action motion synthesis, we create a strong baseline by extending single-action \baselinesingle~\cite{petrovich2021action} approach to multi-actions. To this end, we fix the maximum number of actions to be $M$, and divide latent vector $z$ into $M$ equal-sized sub-vectors each of size $z_i\in\R^{\floor{d/M}}$ to encode one action sub-sequence. We call this approach \baseline. The baseline fails to generalize to Charades dataset such that it synthesizes static frames where the frame appears to match the action. We assume that this is because unlike PROX, the videos in Charades were captured unprofessionally in cluttered scenes and the frames are often blurred. This makes obtaining high-quality pose estimates challenging. As the VAE attempts to smooth out jittery sequences by learning a smooth latent representation, the result is a sequence with hardly any motion. We therefore attempt to find a subset of actions for which the baseline converges to a plausible solution. We found a certain subset of actions of size 30 to yield such a solution, we therefore conduct a direct comparison between our model and the baseline model on this subset. Since our aproach still generalizes to this much more challenging dataset, it demonstrates that is more robust to jitters and noise in motion sequences. 
 
\myparagraph{Standard evaluation metrics.} We follow~\cite{guo2020action2motion} and train an action recognition model~\cite{yu2017spatio} on the GT training data and evaluate on the synthesized sequences using Fréchet Inception Distance \textit{(FID)}, \textit{accuracy} of predicted action labels, \textit{multimodality} and \textit{diversity}, where the latter two refer to the variance between the different actions classes and variance within the same class.

\myparagraph{Semantic consistency metrics.} We evaluate semantic consistency between GT conditioning labels and synthesized sequences. Specifically, we calculate the proportion of sequences whose conditioning labels match the labels of the closest sequence in GT training set. We calculate cross-product distances between each 3D pose in generated sequence and each 3D pose in GT sequence, and find the sequence-to-sequence matching using the Hungarian algorithm.

\subsection{Evaluation on PROX dataset}
\myparagraph{Using standard metrics.} Tab.~\ref{tab:eval_increase_seq_multi} shows the performance of \ours~and comparison to \baseline, while varying sequence length ($60$, $80$, $120$) for sequences with at most $8$ actions. \emph{tr} denotes training set, \emph{gen} refers to synthesized sequences, and \emph{gt} denotes GT data. \ours~significantly improves over \baseline, with a performance gap that rises with an increased sequence length. This is due to the small capacity obtained by the latent vector which does not suffice to represent significant variations in actions when synthesizing multi-action sequences. Tab.~\ref{tab:eval_increase_seq_single} provides direct comparison to~\cite{petrovich2021action} in the single-action use case, with \ourssingle~performing on-par with \baselinesingle~\cite{petrovich2021action}.

\myparagraph{Using semantic consistency metrics.} \ours~significantly outperforms \baseline~when fixing the number of actions to four and varying the sequence length (\ref{fig:all_match_rate} (a)). \ourssingle~matches \baselinesingle~\cite{petrovich2021action} in single-action use cases (\ref{fig:all_match_rate} (b)). Increasing the number of actions for a fixed sequence length degrades~\ours~accuracy much more gracefully compared to~\baseline~(\ref{fig:all_match_rate} (c)).

\begin{figure}[h]
	\setlength\tabcolsep{4pt}
	\resizebox{0.85\textwidth}{!}{%

		\begin{tabular}{llllll}
			{\thead{Getting up from lying position, Standing up}} &{\includegraphics[height=3cm, width=3cm]{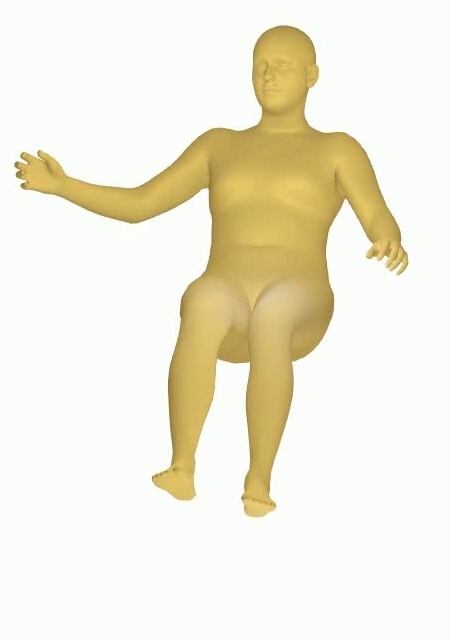}}&{\includegraphics[height=3cm, width=3cm]{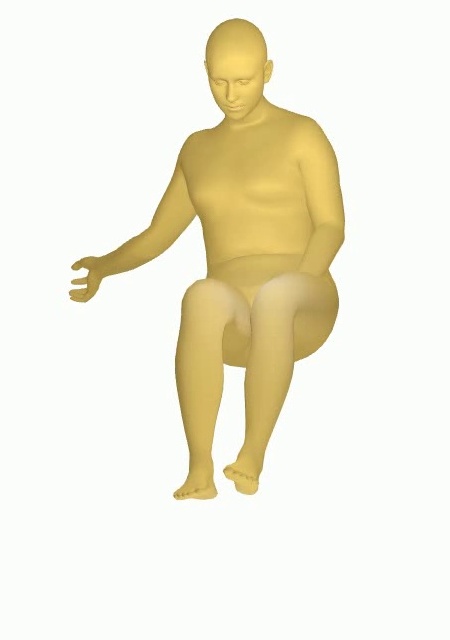}}&{\includegraphics[height=3cm, width=3cm]{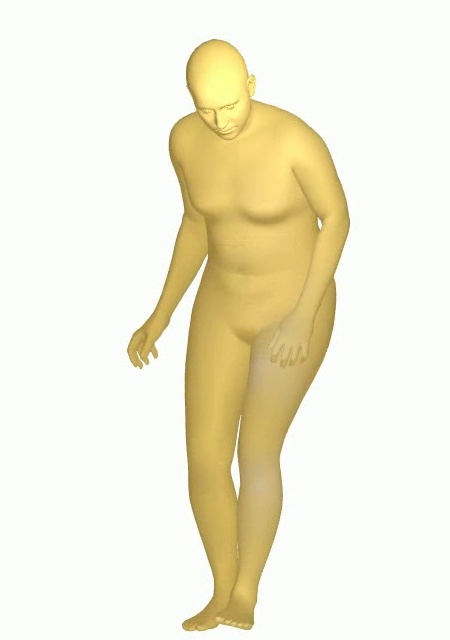}}&  {\includegraphics[height=3cm, width=3cm]{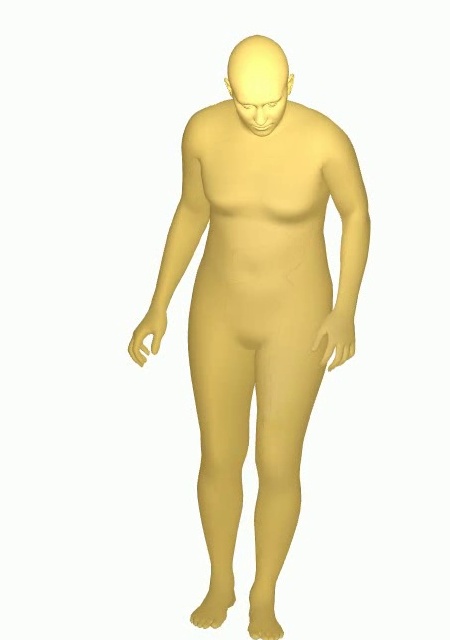}}\\

            {\thead{Reaching for pocket, Punching phone screen} }&{\includegraphics[height=3cm, width=3cm]{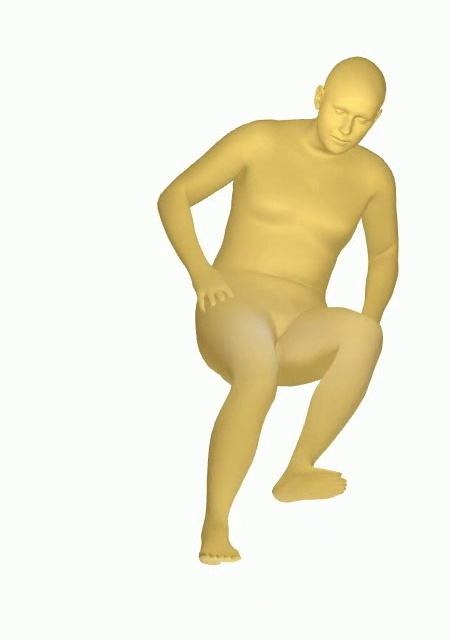}}&{\includegraphics[height=3cm, width=3cm]{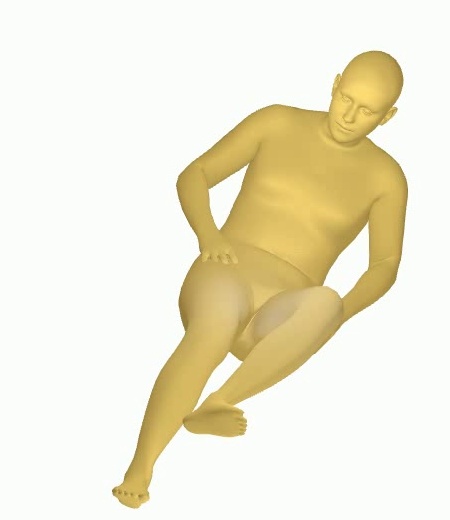}}&{\includegraphics[height=3cm, width=3cm]{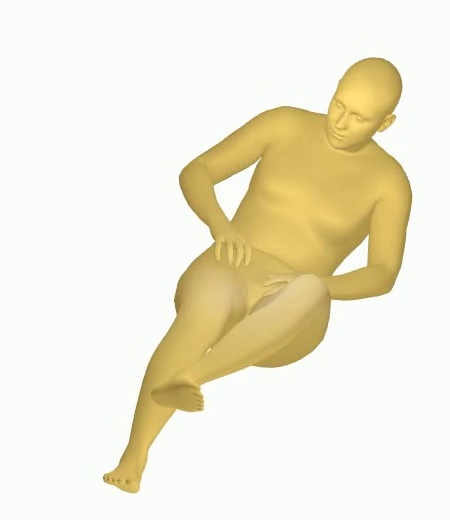}}&  {\includegraphics[height=3cm, width=3cm]{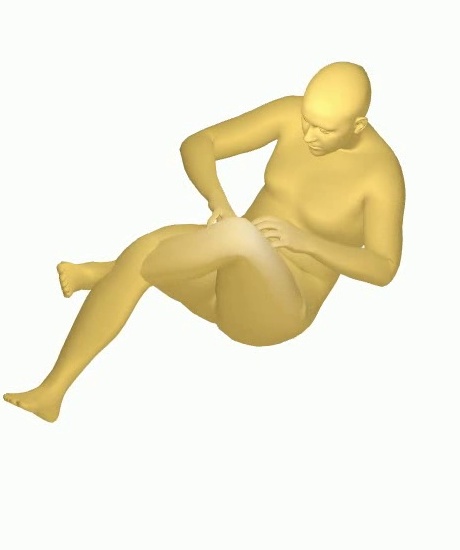}}\\
                        
                        {\thead{Walking, Turning around, Sitting, Reaching for something on table}}&{\includegraphics[height=3cm, width=3cm]{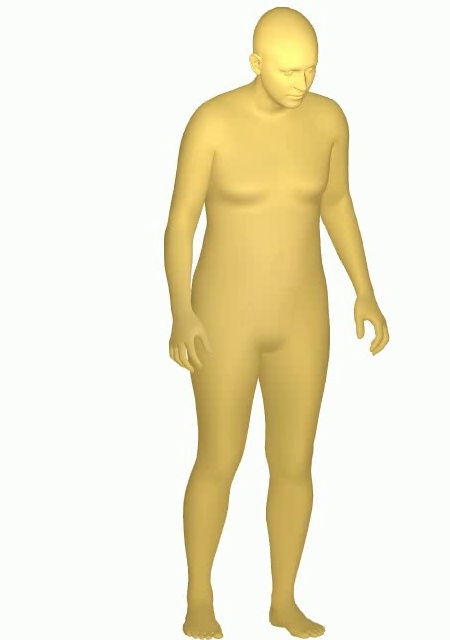}}&{\includegraphics[height=3cm, width=3cm]{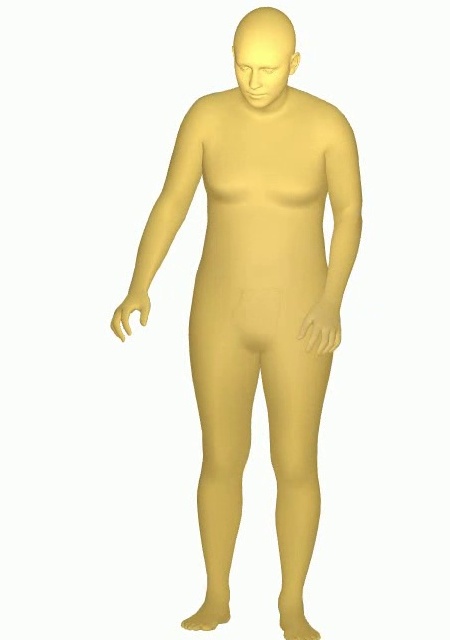}}&{\includegraphics[height=3cm, width=3cm]{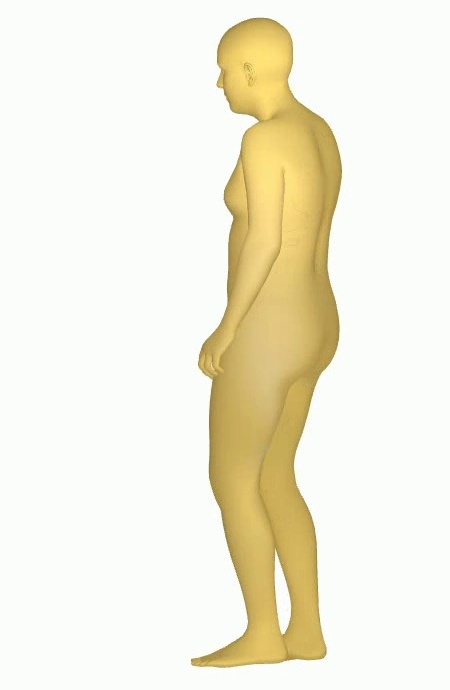}}&  {\includegraphics[height=3cm, width=3cm]{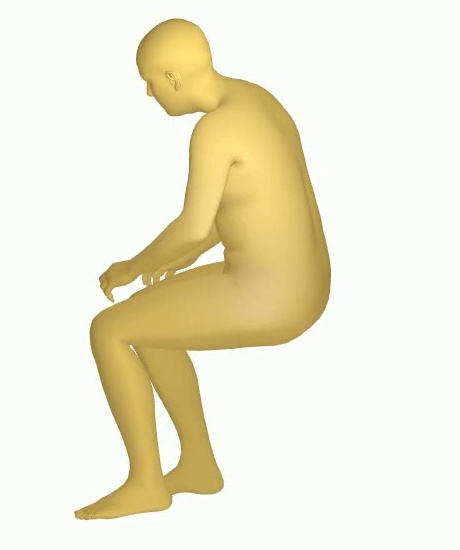}}\\
                         {\thead{Writing on board, Explaining something on board}}&{\includegraphics[height=3cm, width=3cm]{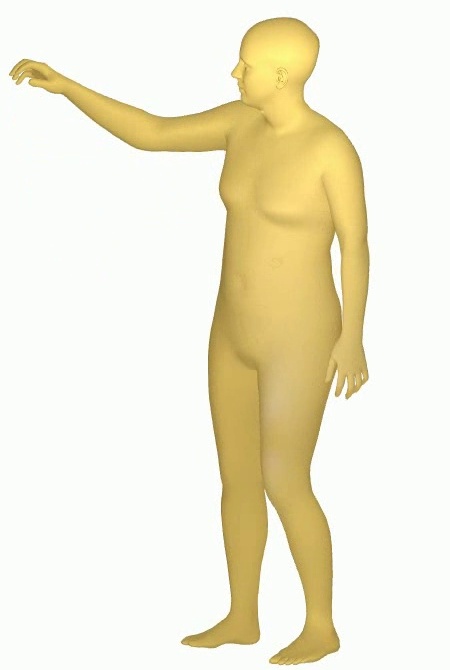}}&{\includegraphics[height=3cm, width=3cm]{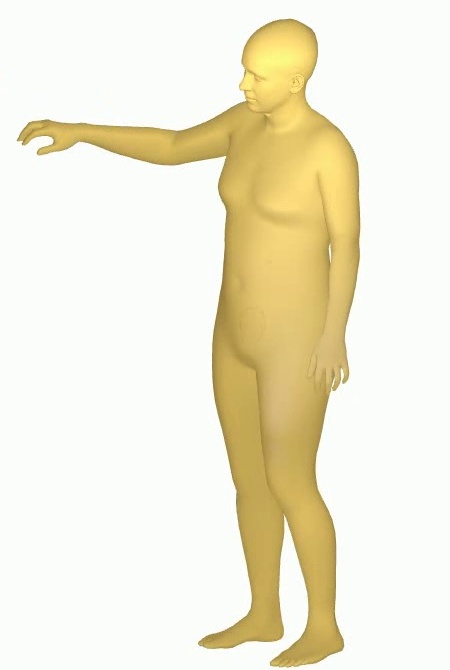}}&{\includegraphics[height=3cm, width=3cm]{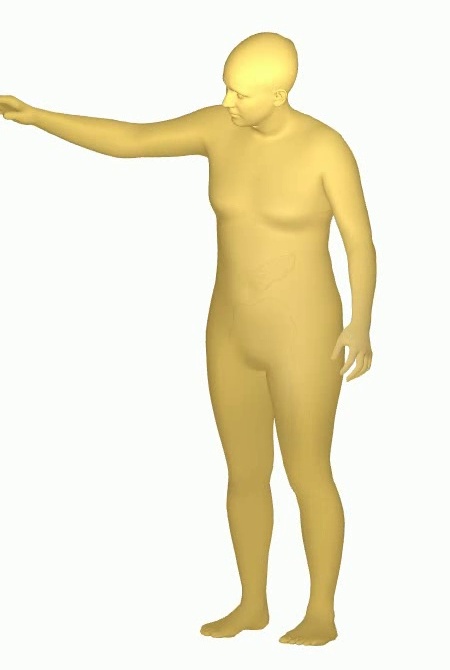}}&  {\includegraphics[height=3cm, width=3cm]{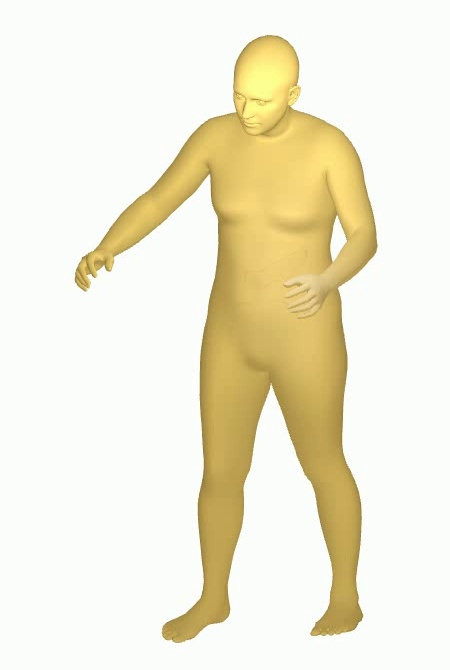}}\\
                                                  {\thead{Lying down, Writing on board}}&{\includegraphics[height=3cm, width=3cm]{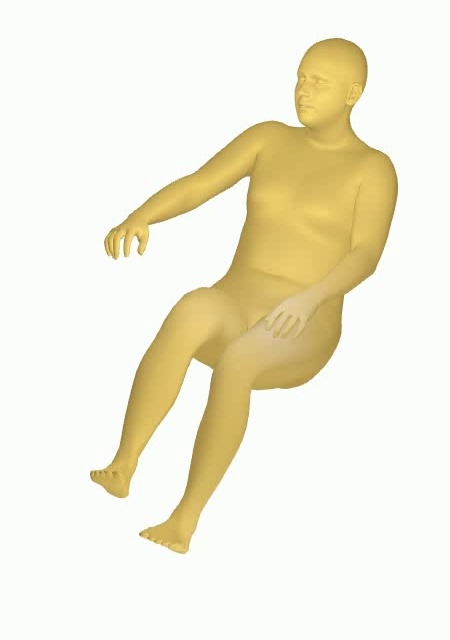}}&{\includegraphics[height=3cm, width=3cm]{images/seq4/frame10}}&{\includegraphics[height=3cm, width=3cm]{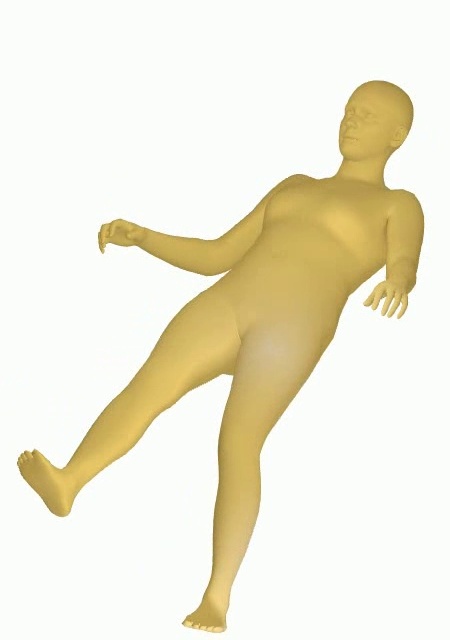}}&  {\includegraphics[height=3cm, width=3cm]{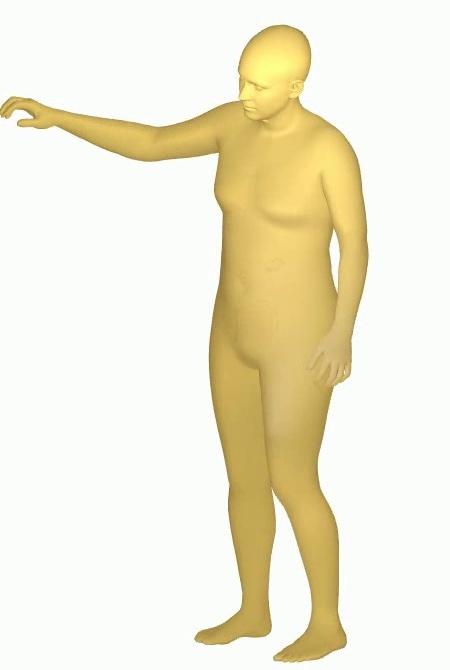}}
                         
	\end{tabular}}
	\caption{Sample action sequences synthesized by \ours~based on PROX dataset. In the last two rows, we feed in two custom actions labels not contained in the training samples. In the fifth row we specify two temporally unrelated actions "sitting down, explaining on board" as conditioning labels and observe that the model struggles to imagine a sequence that satisfies simultaneously both actions, which results in an incoherent sequence. However, in the fourth row, since "writing on board, explaining on board" are temporally and semantically related, we see the synthesized sequence is coherent with a smooth crossover.}

	\label{fig:qual}
\end{figure}

\myparagraph{Qualitative evaluation.} Sample multi-action generated results are shown in Fig. \ref{fig:qual}. 
We observe continuity in the spatial and temporal domain when the actions are temporally related. While our model does not address semantic consistency between subsequent actions, this consistency is inferred based the temporal consistency data distribution which guarantees that generally only related actions may appear close together in the same sequence. 

\begin{figure}[h]
	\setlength\tabcolsep{4pt}
	\resizebox{0.85\textwidth}{!}{%
		\begin{tabular}{llllll}
			{\thead{Putting on shoes, Sitting in a chair, Putting on shoes}} &{\includegraphics[height=3cm, width=3cm]{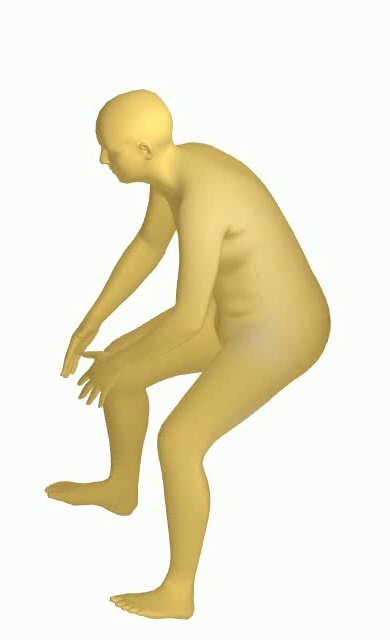}}&{\includegraphics[height=3cm, width=3cm]{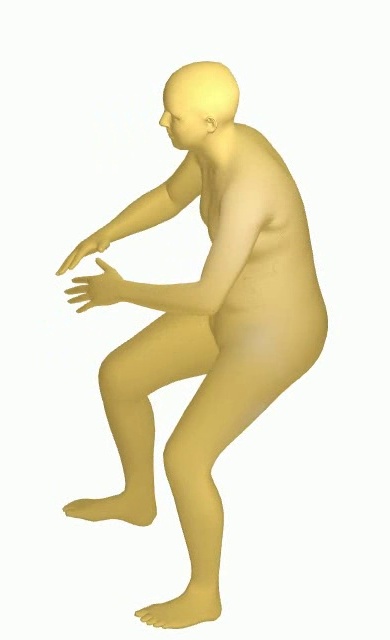}}&{\includegraphics[height=3cm, width=3cm]{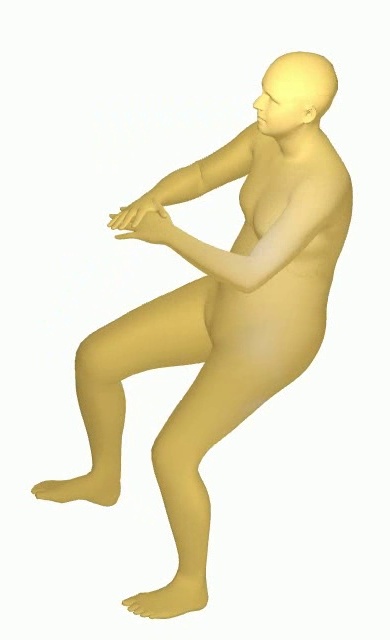}}&{\includegraphics[height=3cm, width=3cm]{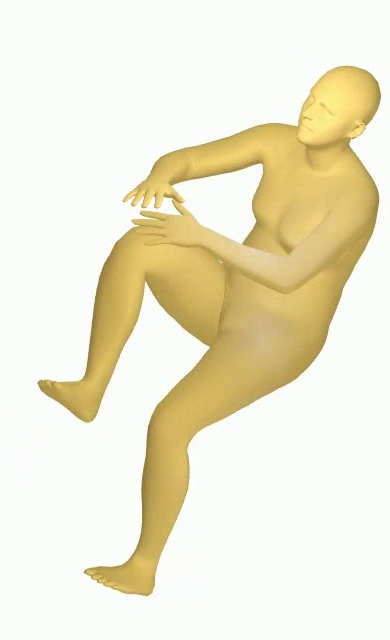}}\\
			
{\thead{Lying on bed, Snuggling with a pillow, Holding a notebook}} &{\includegraphics[height=3cm, width=3cm]{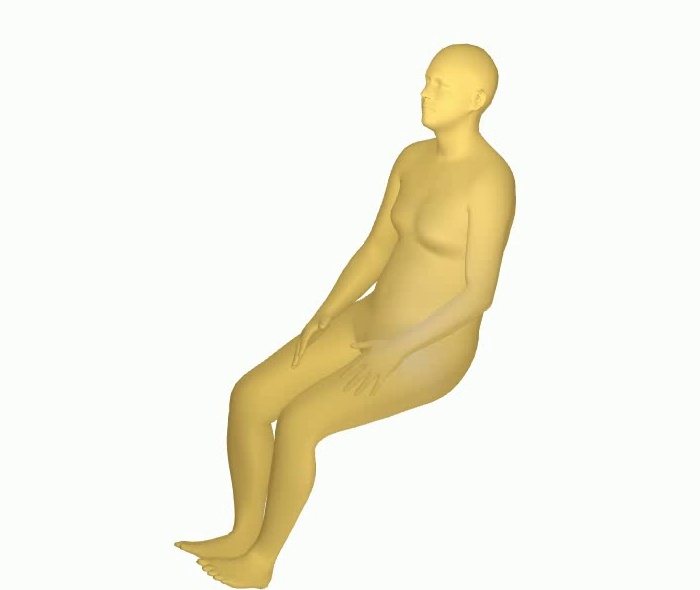}}&{\includegraphics[height=3cm, width=3cm]{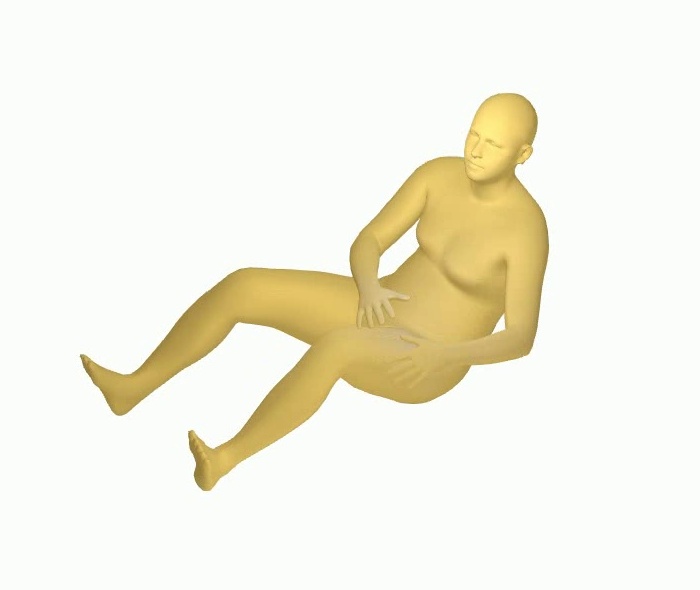}}&{\includegraphics[height=3cm, width=3cm]{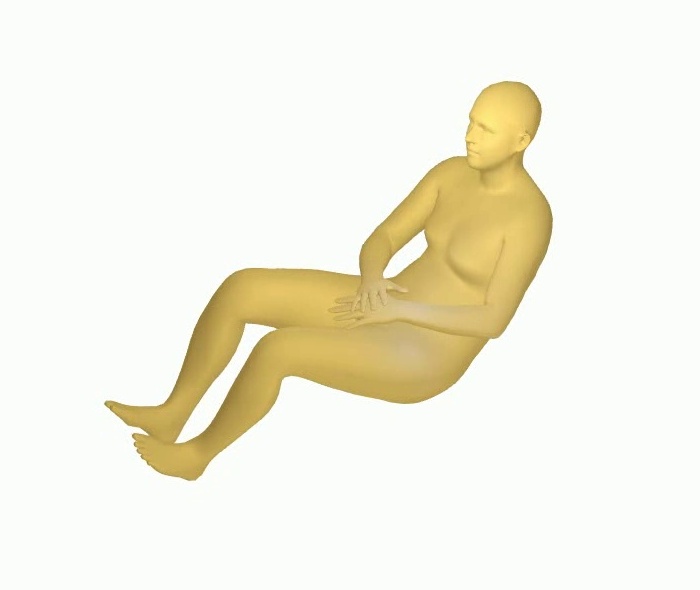}}&  {\includegraphics[height=3cm, width=3cm]{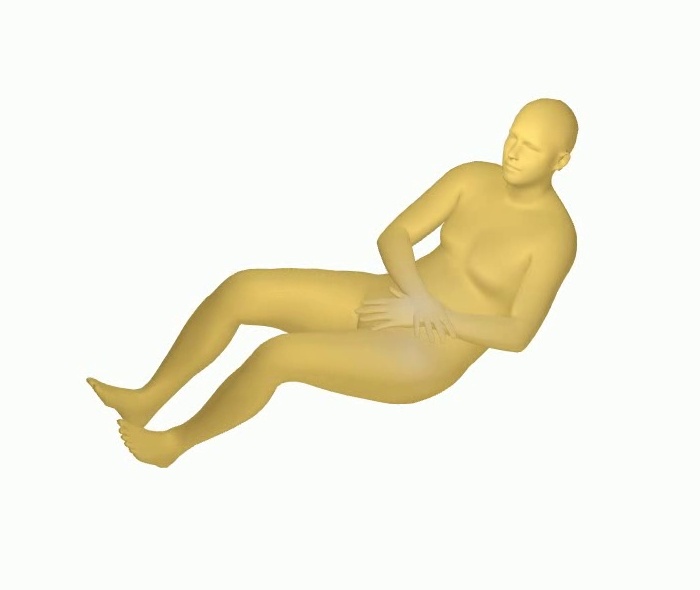}}\\     
          
{\thead{Watching something on laptop, Closing laptop}} &{\includegraphics[height=3cm, width=3cm]{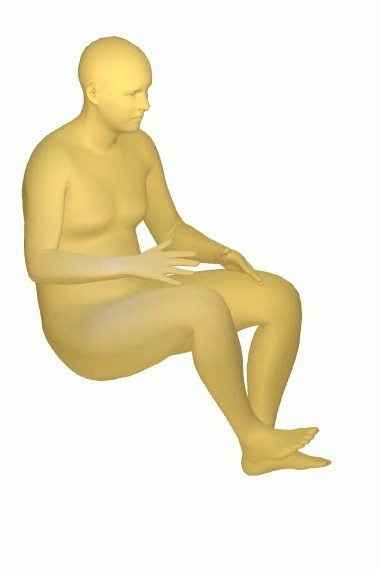}}&{\includegraphics[height=3cm, width=3cm]{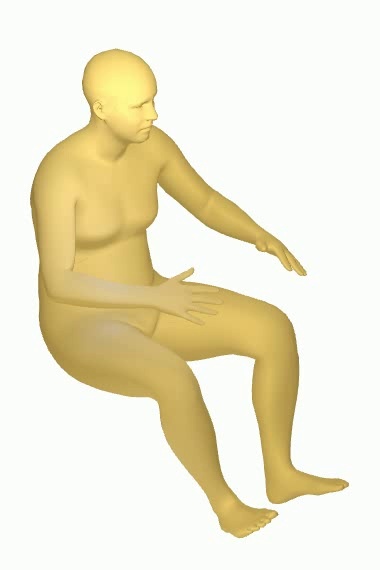}}&{\includegraphics[height=3cm, width=3cm]{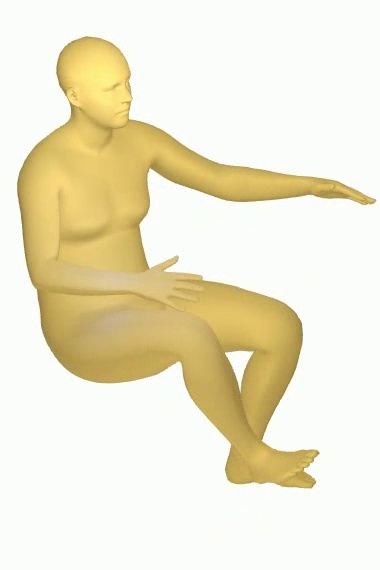}}&  {\includegraphics[height=3cm, width=3cm]{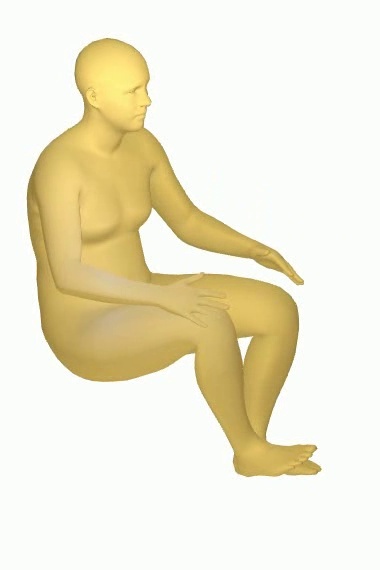}}
	\end{tabular}}
	\caption{Sample action sequences synthesized by \ours~based on Charades dataset.}
	\label{fig:qual_charades}
\end{figure}

\begin{table}[ht]
	\centering
		\begin{adjustbox}{width=0.4\textwidth}
	\small
	\begin{tabular}{lc}
		\hline
		  method &correctly matched sequences \%\\
		  \hline
	        \ours&\bf{78.74}\\
	        Average stats&67.44\\
	        	All diff-latent&67.77\\
	        	Single latent&66.66\\
	        W/o look-b.-a.&28.8
	\end{tabular}
	\end{adjustbox}
	\caption{Ablative evaluations on PROX using \textit{semantic consistency metrics} (max $80$ frames- and max $8$ actions per sequence).}
	\label{tab:eval_matched_labels_ablative}
\end{table}   
\subsection{Evaluation on Charades dataset}
\begin{figure}
\centering
\setlength\tabcolsep{1pt}
\begin{tabular}{cc}
 \includegraphics[width=0.5\linewidth]{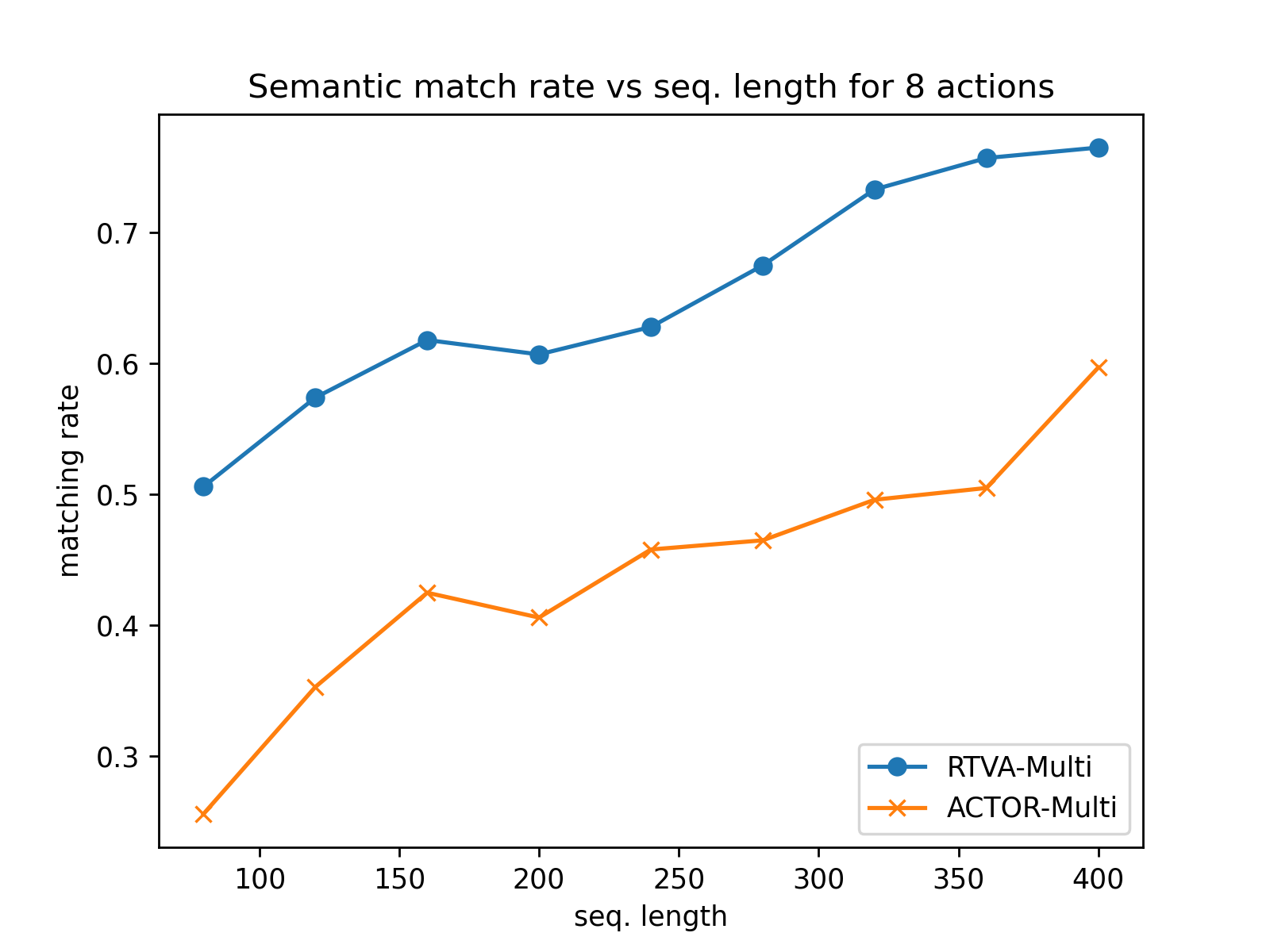}
 &
 \includegraphics[width=0.5\linewidth]{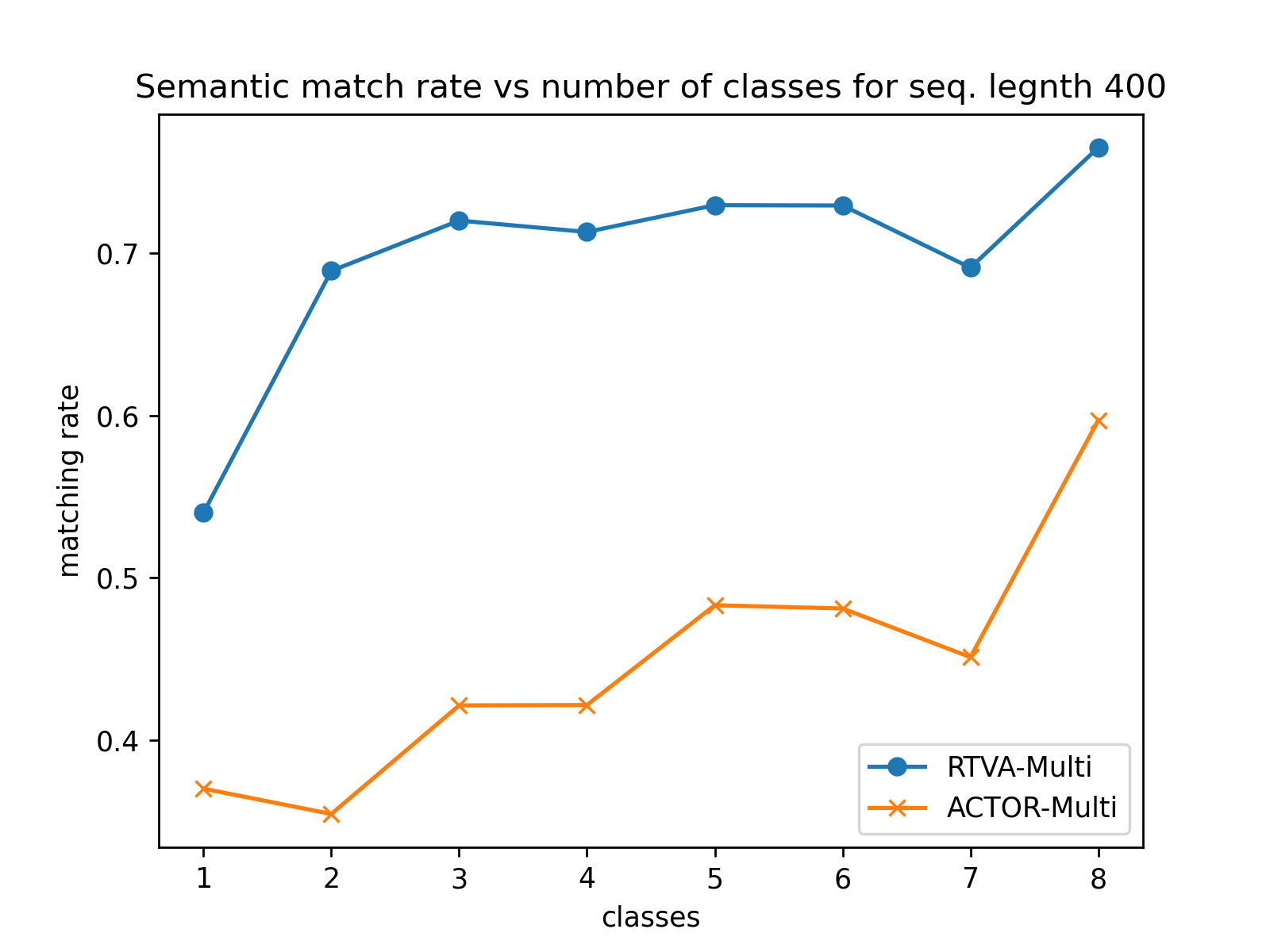}\\
 (a) & (b) 
\end{tabular}
\caption{Evaluation on a subset of 30 actions from Charades using both \baseline and \ours. Match rate of action labels between synthesized and ground-truth samples while varying the sequence length for 8 actions(a) and varying number of actions for a constant sequence length (b)}
\label{fig:match_rate_charades_subset}
\end{figure}

In this section we evaluate our model over Charades dataset. We summarize the results pertaining to standard metrics in Tab.\ref{tab:eval_increase_seq_charades}. Similarly, we also report the results of semantic consistency in Fig. \ref{fig:match_rate_charades_subset}.
\begin{table*}[ht]
	\centering
	\begin{adjustbox}{width=1.0\textwidth}
	\small
	\begin{tabular}{lccccccccccc}
		\hline
		  seq. length&method &accuracy\textsubscript{tr\textsubscript{gt}}& accuracy\textsubscript{tr\textsubscript{gen}}$\uparrow$&FID\textsubscript{tr\textsubscript{gen}}$\downarrow$&diversity\textsubscript{tr\textsubscript{gt}}&multimodality\textsubscript{tr\textsubscript{gt}}&diversity\textsubscript{tr\textsubscript{gen}}&multimodality\textsubscript{tr\textsubscript{gen}}\\
		  \hline
		  \multirow{2}{*}{60} &\baseline&\multirow{2}{*}{$71.6$}& $22.3\pm0.87$&$\textbf{909.99}\pm 5.66$&\multirow{2}{*}{$73.78$}&\multirow{2}{*}{$45.32$}&{$46.11\pm 0.42$}&$36.22\pm0.51$\\
		&\ours&&$\textbf{48.7}\pm0.77$&$1039.82\pm7.40$&&&$50.33\pm0.53$&$29.8\pm0.35$\\
		\hline	
		\multirow{2}{*}{80} &\baseline&\multirow{2}{*}{$84.3$}& $28.5\pm0.97$&$\textbf{1177.74}\pm 11.72$&\multirow{2}{*}{$86.57$}&\multirow{2}{*}{$26.96$}&{$82.95\pm 0.6$}&$55.33\pm0.54$\\
		&\ours&&$\textbf{54.3}\pm0.40$&$1549.12\pm9.66$&&&$56.74\pm0.52$&$36.53\pm0.45$\\
		\hline
		\multirow{2}{*}{120} &\baseline&\multirow{2}{*}{$77$}&$24.7\pm0.60$&$\textbf{2845.65}\pm 23.42$&\multirow{2}{*}{118.19}&\multirow{2}{*}{83.64}&$75.84\pm0.76$&$59.76\pm0.74$\\
		&\ours& &$\textbf{38.7}\pm0.52$&$3593.27\pm31.2$&&&$73.47\pm0.73$&$49.76\pm0.64$\\
		
	\end{tabular}
	\end{adjustbox}
	\caption{Evaluation on a subset of 30 actions from Charades dataset with a maximum of $8$ actions/sequence, while varying the sequence length.} 
	\label{tab:eval_increase_seq_charades}
\end{table*} 

\subsection{Ablative studies}
Ablative studies performed using \textit{standard metrics} (Tab.~\ref{tab:eval_ablative}) and \textit{semantic consistency metrics} (Tab.~\ref{tab:eval_matched_labels_ablative}) on PROX dataset. 

\myparagraph{Last frame distribution parameters.} \ours~uses distribution parameters of the last frame of each action label. We evaluate the performance when using averaged parameters over all frames in an action sub-sequence. We observe deteriorated performance in this case, indicating that the learned distribution becomes better as the model accumulates more context information about the sequence (see \textit{Average stats} in Tab. \ref{tab:eval_ablative} and \ref{tab:eval_matched_labels_ablative}).

\myparagraph{Different latent vector per frame.} Using different latent vectors for each generated frame also adversely affects performance. This indicates that a single latent vector representing the whole action sub-sequence provides a smoother frame generation, as nearby frames should be close together in the latent space (see \textit{All diff-latent} in Tab. \ref{tab:eval_ablative} and \ref{tab:eval_matched_labels_ablative}).  

\myparagraph{Single latent vector per sequence.} We create $k$ latent vectors where we sample only a single latent vector and add each of the $k$ action embeddings to one instance. This results in an over-smoothed sequence such that transitioning to a new action barely happens. (see \textit{Single latent} in Tab. \ref{tab:eval_ablative} and \ref{tab:eval_matched_labels_ablative}). 

 \myparagraph{Look-back-ahead.} Stacking all latent vectors during the decoding phase is necessary for the synthesized sequence to avoid discontinuity between different action sub-sequences. We call it \textit{look-back-ahead} strategy as it makes the decoder aware of both previous and future conditioning actions while reasoning about current frame. Using a separate latent vector from only the current action results in a dramatic performance drop (see \textit{W/o look-b.-a.} in  Tab.~\ref{tab:eval_ablative} and \ref{tab:eval_matched_labels_ablative}).

\section{Conclusion}
We addressed a challenging problem of synthesizing multi-action human motion sequences. We proposed a novel spatio-temporal formulation combining the expressiveness and efficiency of Recurrent Transformers with generative richness of conditional VAEs. The proposed approach generalizes to an arbitrary number of actions and frames per generated sequence, with space and time requirements growing linearly in the number of frames. Experimental evaluation showed significant improvements in multi-action motion synthesis over the state-of-the-art.

\medskip

{\small
\bibliographystyle{ieee_fullname}
\bibliography{egbib}
}

\end{document}